\pdfoutput=1

\documentclass[11pt]{article}

\usepackage[]{emnlp2021}

\usepackage{times}
\usepackage{url}
\usepackage{latexsym}
\usepackage{amssymb}

\usepackage{titlesec}
\usepackage{multirow}
\usepackage{subfig}
\usepackage{mathtools}
\usepackage{colortbl}
\usepackage{float}
\usepackage{xcolor}
\usepackage{enumitem}
\usepackage[capitalize]{cleveref}

\usepackage{booktabs}
\usepackage[font=small,labelfont=bf,tableposition=top]{caption}
\DeclareCaptionLabelFormat{andtable}{#1~#2  \&  \tablename~\thetable}

\definecolor{intro_green}{RGB}{84, 130, 53}
\definecolor{intro_purple}{RGB}{112, 48, 160}
\definecolor{intro_orange}{RGB}{237, 125, 49}

\usepackage{tikz}

\usepackage[T1]{fontenc}

\usepackage[utf8]{inputenc}

\usepackage{microtype}

\title{Context-Aware Interaction Network for Question Matching}

\author{Zhe Hu$^{1}$, Zuohui Fu$^{2}$, Yu Yin$^{3}$, and  Gerard de Melo$^{4}$
\\
  $^{1}$Baidu Inc., China,
  $^{2}$Rutgers University, USA \\
  $^{3}$Northeastern University, USA,
  $^{4}$HPI/University of Potsdam, Germany \\
$^{1}${\tt huzhe01@baidu.com},
  $^{2}${\tt zuohui.fu@rutgers.edu} \\
  $^{3}${\tt yin.yu1@northeastern.edu},
  $^{4}${\tt gdm@demelo.org}
  }

\begin{document}
\maketitle

\begin{abstract}
Impressive milestones have been achieved in text matching by adopting a cross-attention mechanism to capture pertinent semantic connections between two sentence representations.
However, regular cross-attention focuses on word-level links between the two input sequences, neglecting the importance of contextual information.
We propose a context-aware interaction network (COIN) to properly align two sequences and infer their semantic relationship.
Specifically, each interaction block includes (1) a context-aware cross-attention mechanism to effectively integrate contextual information when aligning two sequences, and (2) a gate fusion layer to flexibly interpolate aligned representations.
We apply multiple stacked interaction blocks to produce alignments at different levels and gradually refine the attention results. 
Experiments on two question matching datasets and detailed analyses demonstrate the effectiveness of our model.

\end{abstract}

\section{Introduction}
Semantic text matching is among the most fundamental tasks in natural language processing. Given two sentences, the goal is to predict their semantic relationship. In this work, we focus in particular on question matching (QM) benchmarks.

Recently, the availability of large-scale annotated datasets has led to a proliferation of deep neural architectures for text matching~\cite{williams-etal-2018-broad,chen-etal-2017-enhanced,wang2017bilateral}. Most existing neural models fall into two categories, namely the sentence encoding and the sentence interaction approaches~\cite{lan-xu-2018-neural}. The former encodes sentences as fixed-length vector representations, which are then consulted to make the final prediction. The latter considers interactions between two sequences to identify their semantic connections, which tends to yield better results.

\begin{figure}[t]
    \centering
    \includegraphics[scale=0.38]{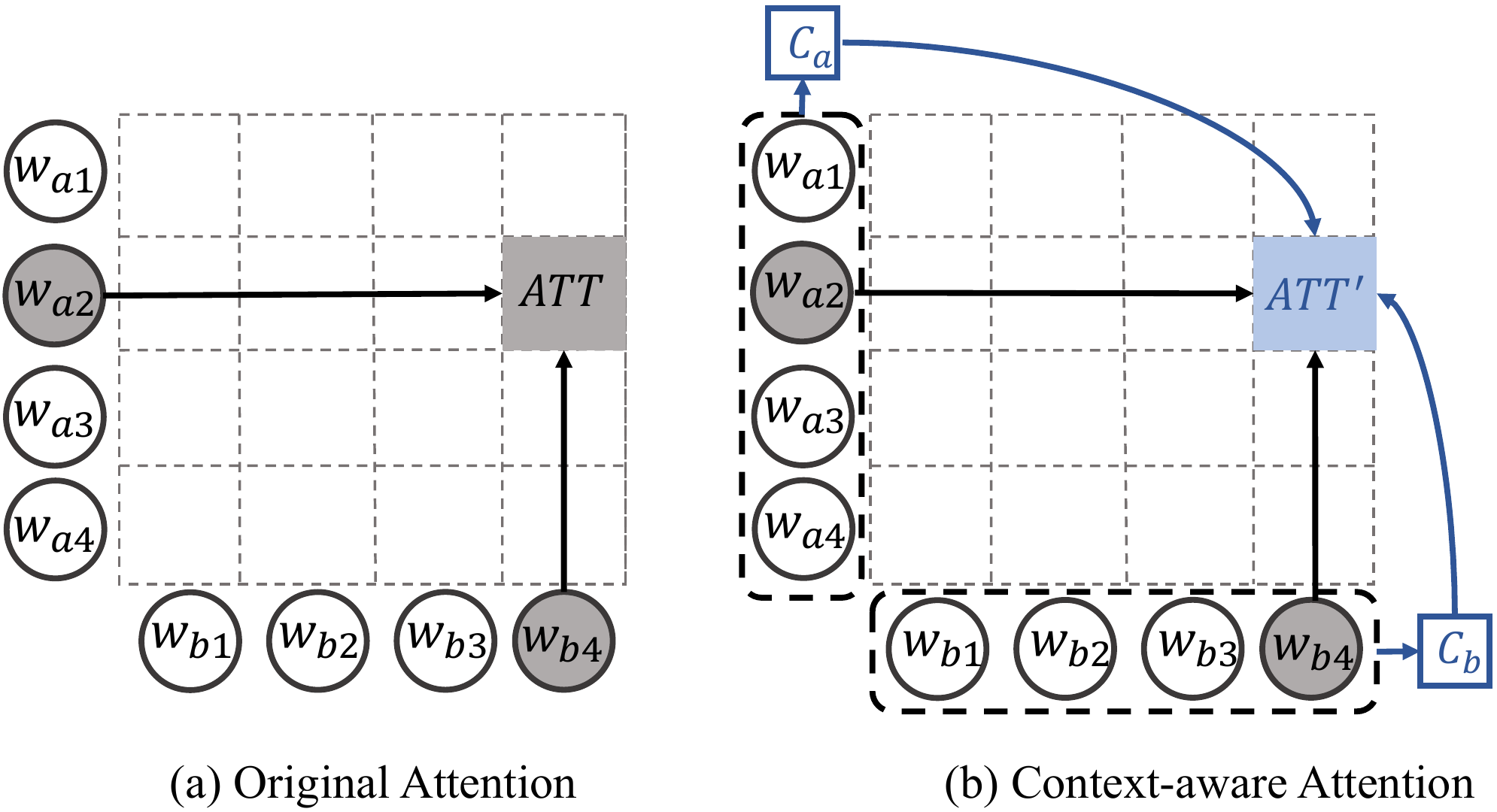}
    \vspace{-2mm}
    \captionof{figure}{ 
    The original attention mechanism (left) and the proposed context-aware attention (right). 
    $w_{*}$ represents the two sequences (more generally, they can be regarded as query and key). $C_{*}$ denotes the contextual features.
    }
    \label{fig:att_descript}
    \vspace{-4mm}
\end{figure}

Attention mechanisms are widely adopted for the sentence-interaction approaches, relying on a word-by-word attention matrix to obtain alignment information between two sequences. This has proven fruitful in modeling sentence pair relationships~\cite{parikh-etal-2016-decomposable,rocktaschel2015reasoning,wang-jiang-2016-learning}.
Nonetheless, when computing the cross-sentence attention, existing models mostly focus on word-level local matching and fail to fully account for 
the overall semantics: \textit{each value of the attention matrix is based on just two individual tokens from the sequences without full consideration of the context}. 
As shown in Figure~\ref{fig:att_descript}, in the original attention mechanism, each token individually attends to the other tokens without accounting for important contextual information. 
However, accurate matching may require a deeper understanding of the two sentences along with pertinent linguistic patterns and constructions~\cite{storks2019recent}.
\newcite{yang2019context} show that 
contextualizing the self-attention network may improve the original representations, but they do not consider the scenario of 
sentence pairs with cross-attention.

In this work, we aim to generalize the notion of cross-sentence attention by enabling it to incorporate rich contextual signals. We propose a 
\underline{CO}ntext-aware \underline{I}nteraction \underline{N}etwork
(\textbf{COIN}) with a novel \textit{context-aware attention layer}. 
This layer enables the model to consult contextual information while computing the cross-attention matrix to measure the word relevance, yielding
better contextualized alignments for semantic reasoning.
We leverage the self-alignment on each sequence to produce contexts that represent salient features for each token. The subsequent gate fusion layer is designed to enable the model to selectively integrate the aligned representations and control to what extent the new information is to be passed to the following layers, which is similar to a skip connection in mitigating the additional model complexity coming from the deeper structure. Finally, an aggregation layer and a multi-head pooling layer are adopted to infer high-level semantic representations for the sequences and predict the result based on the refined representations.

To validate the effectiveness of our method, we conduct extensive experiments on the Quora and LCQMC datasets, along with further analyses of model components and a case study visualizing the alignment. The results show that by incorporating rich context into cross-attention, our model outperforms state-of-the-art methods without the huge number of model parameters and pre-training on extrinsic data of BERT models.


\section{Method}

\begin{figure}[t]
    \centering
    \includegraphics[scale=0.58]{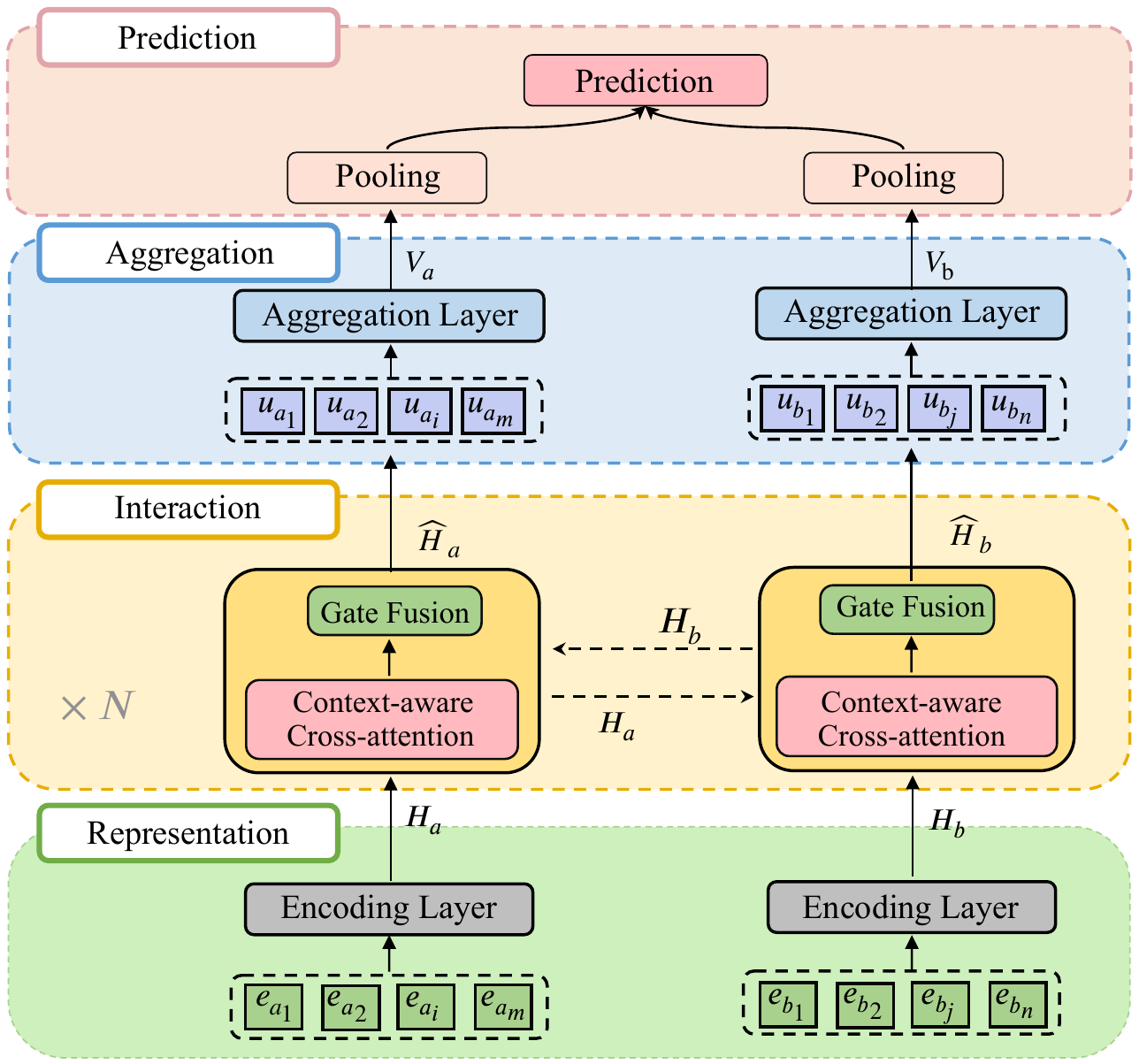}
    \captionof{figure}{ 
    Overview of our model structure.
    }
    \label{fig:model_structure}
\end{figure}

Question matching can be viewed as a classification task that seeks a label $y \in \mathcal{Y}= \{\textsc{Duplicate}, \textsc{Non-Duplicate}\}$ for a given sentence pair $(S_a, S_b)$. Figure~\ref{fig:model_structure} illustrates our novel sentence interaction approach for this task. In the following, we describe the individual ingredients of this approach.

\subsection{Input Representation Layer}
The input representation layer converts each sentence into matrix representations with an embedding and encoding layer. We invoke word embeddings without additional lexical features and adopt a multi-layer convolutional encoder on top of the embedding layer. In addition, we concatenate the contextual representations with the original embeddings to produce better alignments in the following interaction blocks. This serves a similar purpose as skip connections to represent words at different levels~\cite{wang-etal-2018-multi}.

\subsection{Context-aware Interaction Block}\label{Self-aware Interaction Block}
Our proposed interaction block consists of a context-aware cross-attention and a gate fusion layer. Several such interaction blocks are stacked to obtain refined alignments.
\subsubsection{Cross-Attention Layer} 
We first review the original cross-sentence attention before introducing our context-aware form of attention. Assume the two inputs of the current layer are
$\mathbf{H}_a=(\mathbf{h}_{a_1}, ..., \mathbf{h}_{a_m})$ 
and 
$\mathbf{H}_b=(\mathbf{h}_{b_1}, ..., \mathbf{h}_{b_n})$, 
where $m$ and $n$ are the corresponding sequence lengths. The word-by-word attention matrix is first calculated as follows: 

\vspace{-4mm}
{\fontsize{10}{11}\selectfont

\begin{align}
&\mathbf E_{ij} = {\rm Att}(\mathbf{h}_{a_i}, \mathbf{h}_{b_j}) = F_{1}(\mathbf{h}_{a_i})^\mathrm{T} F_{1}(\mathbf{h}_{b_j}), \label{eq:attention}\vspace{-5mm}
\end{align}%
}%
\noindent where $F_1$ is a feed-forward neural network. Then the similarity matrix $\mathbf E$ is used to compute aligned representations of each sequence as a weighted summation with regard to the other sentence:

\vspace{-4mm}
{\fontsize{10}{11}\selectfont
\begin{align}
&\mathbf a_{i} = \rm softmax(\mathbf E_{i:}),\quad \mathbf b_{j} = \rm softmax(\mathbf E_{:j}) \\
& \mathbf{h}^{\prime}_{b_j} = {\sum_{k=1}^{m}}{\mathbf b_{kj} \mathbf h_{a_k}},\quad \mathbf{h}^{\prime}_{a_i} = {\sum_{k=1}^{n}}{\mathbf a_{ik} \mathbf h_{b_k}}\vspace{-5mm}
\end{align}%
}\vspace{-4mm}

\noindent \textbf{Limitation.} It is evident in Eq.~\ref{eq:attention} that each value of the attention matrix is governed by the parameters of the feed-forward layer with respect to only the individual token pairs, 
so the layer does not take advantage of valuable contextual signals.

\subsubsection{Context-Aware Cross-Attention Layer} 
We propose a novel context-aware cross-attention layer by incorporating  contextual representations into the cross-attention. The goal is to enable the model to identify salient contextual features for each token, and consider these features when computing the cross-attention matrix $\mathbf E$. 

Given $\mathbf C_{a}= (\mathbf c_{a_1}, ..., \mathbf c_{a_m})$, $\mathbf C_{b}=(\mathbf c_{b_1}, ..., \mathbf c_{b_n})$ as contextual representations for the two sentences, we modify the attention mechanism from Eq.~\ref{eq:attention} to be able to draw on these as additional inputs when computing the word-by-word attention matrix:

\vspace{-2mm}
{\fontsize{10}{11}\selectfont
\begin{align}
\begin{split}
\mathbf E^\mathrm{c}_{ij} &= {\rm Att_{context}}(\mathbf{h}_{a_i}, \mathbf{h}_{b_j}, \mathbf{c}_{a_i}, \mathbf{c}_{b_j}) \\
&= F_{1}(\mathbf{h}_{a_i} + \mathbf{c}_{a_i} )^\mathrm{T} F_{1}(\mathbf{h}_{b_j} + \mathbf{c}_{b_j})
\end{split}\vspace{-1mm}
\end{align}%
}%

\noindent By incorporating the contextual vectors, the model is able to take advantage of the full context and enable better alignments. 

\smallskip
\noindent \textbf{Contextual Representations.} In order to compute such representations of the contexts, given each sequence, we adopt a self-alignment layer to aggregate pertinent contextual information. Each contextual vector is computed by attending to the input hidden states and conducting a weighted summation. Formally, for the input $\mathbf H = (\mathbf{h}_1, ..., \mathbf{h}_n)$:

\vspace{-2mm}
{\fontsize{10}{11}\selectfont
\begin{align}
&\mathbf A = {\rm ReLU}(\mathbf W_c\mathbf H)^\mathrm{T}\, {\rm ReLU}(\mathbf W_c\mathbf H) \\
&\mathbf C = \rm softmax(\mathbf A) \mathbf H
\end{align}
\vspace{-4mm}
}

\noindent Here, $\mathbf{W}_c$ is a trainable parameter.

Leveraging the self-alignment to produce contextual signals also mirrors human behavior in the sense that when matching two sentences, people tend to first process each sentence paying attention to the important contents, and then compare the two sentences and connect relevant elements (words or phrases) with contextual features to identify their relationship, rather than just comparing individual words.

\subsubsection{Gate Fusion Layer} 
Subsequently, a gate fusion layer compares the original sequences against the aligned representations and blends them together as new sequence representations. 
Specifically, we first compare the original representation ($\mathbf{H}_{a}$) with the aligned one ($\mathbf{H}^{\prime}_{a}$) from three perspectives, and then combine them with a non-linear transformation:

\vspace{-5mm}
{\fontsize{10}{11}\selectfont

\begin{align}
&{\mathbf{\widetilde{h}}}^{1}_{a_i} = G_1([\mathbf{h}_{a_i}; \mathbf{h}^{\prime}_{a_i}]) \label{eq:ori}\\
&{\mathbf{\widetilde{h}}}^{2}_{a_i} = G_2([\mathbf{h}_{a_i}; \mathbf{h}_{a_i} - \mathbf{h}^{\prime}_{a_i}]) \\
&{\mathbf{\widetilde{h}}}^{3}_{a_i} = G_3([\mathbf{h}_{a_i}; \mathbf{h}_{a_i} \odot \mathbf{h}^{\prime}_{a_i}]) \\
&{\mathbf{\widetilde{h}}}_{a_i} = \mathrm{ReLU}(\mathbf{W}_f[{\mathbf{\widetilde{h}}}^{1}_{a_i};{\mathbf{\widetilde{h}}}^{2}_{a_i};{\mathbf{\widetilde{h}}}^{3}_{a_i}] + \mathbf{b}_f)
\label{eq:sim}
\end{align}
\vspace{-6mm}
}

\noindent 
Then a gated connection is applied to enable the model to selectively integrate the aligned features:
\newcommand{\vect}[1]{\boldsymbol{#1}}
\vspace{-6mm}
{\fontsize{10}{11}\selectfont

\begin{align}
& \vect{f}_i = \sigma(
\mathbf{W}_{1}\mathbf{h}_{a_i} +
\mathbf{W}_{2}{\mathbf{\widetilde{h}}}_{a_i} +
\mathbf{b}_g
) \\
&\widehat{\mathbf{h}}_{a_{i}} = \vect{f}_i \odot \mathbf{h}_{a_i} +
(\mathbf1 - \vect{f}_i) \odot {\mathbf{\widetilde{h}}}_{a_i}
\end{align}
}
\vspace{-6mm}

\noindent Here $\sigma$ is a Sigmoid nonlinear transformation, while $\mathbf{W}_{\ast}$ and $\mathbf{b}_g$ are trainable parameters. The same operation is conducted on sentence $S_b$, thereby yielding the outputs $\widehat{\mathbf{H}}_{a}$ and $\widehat{\mathbf{H}}_{b}$. With these operations, the model can flexibly interpolate the aligned information by controlling the gate, especially when multiple interactions are applied.

\subsection{Aggregation Layer}\label{Aggregation Layer}
To obtain high-level semantic representations for each sentence, we apply another convolutional neural network on top of the interaction blocks to obtain the aggregated sentence representations $\mathbf{V}_a$, $\mathbf{V}_b$, serving as the inputs for the prediction layer.

\subsection{Pooling and Prediction Layer}\label{Prediction Layer}
We compute a weighted summation of the hidden states to get sentence vectors.
To allow the model to represent each sequence in different representation subspaces, we adopt multi-head pooling following \newcite{liu-lapata-2019-hierarchical}. For each head $z$, we first transform the sequence into attention scores $\mathbf S^z$ and values $\mathbf{\widetilde V}^z$:

\vspace{-6mm}
{\fontsize{10}{11}\selectfont

\begin{align}
& \mathbf S^z = {\rm softmax}(\mathbf W^z_a\mathbf V_a) \\
& \mathbf{\widetilde V}^z = \mathbf{W}^z_v\mathbf{V}_{a}
\end{align}
\vspace{-6mm}
}

\noindent where $\mathbf W^z_a \in \mathbb{R}^{1 \times d}$ and $\mathbf W^z_u \in \mathbb{R}^{d_\mathrm{h} \times d}$ are trainable parameters,  with $d_\mathrm{h} = d / n_\mathrm{h}$ as the dimensionality of each head and $n_\mathrm{h}$ as the number of heads. The pooling vector of head $z$ is
computed as

\vspace{-4mm}
{\fontsize{10}{11}\selectfont
\begin{align}
& {\mathbf{\widehat V}^z} = {\sum_{i=1}^{n}}{{\mathbf s^z_i}{\mathbf {\widetilde v}^z_i}},
\end{align}
\vspace{-4mm}
}

\noindent where $\mathbf s^z_i$ and $\mathbf {\widetilde v}^z_i$ denote the calculated attention scores and values. 
The pooling vectors of all heads are  concatenated to form the final vector representations of each sequence $\mathbf{V}^{\prime}_{a}$ and $\mathbf{V}^{\prime}_{b}$. We combine $\mathbf{V}^{\prime}_{a}$ and $\mathbf{V}^{\prime}_{b}$ to produce the overall representation by concatenating the different operations:

\vspace{-4mm}
{\fontsize{10}{11}\selectfont

\begin{align}
&\mathbf{V}=[\mathbf{V}^{\prime}_{a};\mathbf{V}^{\prime}_{b};
\mathbf{V}^{\prime}_{a} - \mathbf{V}^{\prime}_{b}; \mathbf{V}^{\prime}_{a} \odot \mathbf{V}^{\prime}_{b}]
\end{align}
\vspace{-4mm}
}

\noindent Finally, the prediction layer takes the representation $\mathbf{V}$ and passes it to a fully-connected network component to predict the ultimate target scores.

\section{Experiments}

\subsection{Experimental Setup}
\noindent\textbf{Datasets.} 
We conduct experiments on two datasets: 1) The Quora Questions Pairs corpus (Quora) contains over 400k English question pairs selected from Quora.com, for which we use the same data split as \newcite{wang2017bilateral}. 2) LCQMC~\cite{liu-etal-2018-lcqmc} is a large-scale open-domain Chinese question matching corpus constructed from Baidu Knows. We follow the data splits in the original papers, and apply a hard cut-off of the sentence length on both datasets by cropping or padding. The length is set to 32 for Quora and 50 for LCQMC.

\smallskip
\noindent\textbf{Training Details and Parameters.} 
For Quora, we use 300 dimensional GloVe embeddings~\cite{pennington-etal-2014-glove}.
For LCQMC, following \newcite{li-etal-2019-word-segmentation}, we avoid word segmentation and instead use a randomly initialized character embedding matrix. The kernel size is 3 for convolutional layers with padding. We tune the dimensionality of the feed-forward layers from 150 to 300. The batch size is tuned from 32 to 128. Adam optimization is used with an initial learning rate of 0.001 and exponential decay. We 
use ReLU~\cite{pmlr-v15-glorot11a} as the activation function in all feed-forward networks. To prevent over-fitting, dropout with a retention probability of 0.8 is applied. We apply 3 context-aware interaction blocks for Quora and 2 interaction blocks for LCQMC. For BERT~\cite{devlin-etal-2019-bert}, we choose the BERT-base version (12 layers, 768 hidden dimensions and 12 attention heads). Further training details are given in the appendix.

\begin{table}[t]
\fontsize{8}{10}\selectfont
 \setlength{\tabcolsep}{1.56mm}
  \centering
    \begin{tabular}{|l|c|c|}
        \hline
        {\bf Model} & {\bf  Acc (\%)} & {\bf F1 (\%)} \\
        \hline
        
        Lattice-CNN & 82.1 & 82.4 \\
        ESIM~\cite{chen-etal-2017-enhanced} & 82.0 & 84.0 \\
        BiMPM~\cite{wang2017bilateral} & 83.3 & 84.9 \\
        GMN~\cite{chen-etal-2020-neural-graph} & 84.6 & 86.0 \\
        COIN (Ours)  &  {\bf85.6}& {\bf 86.5} \\
        \hline
        BERT~\cite{devlin-etal-2019-bert} & 85.7 & 86.8 \\
        SBERT~\cite{reimers-gurevych-2019-sentence} & 85.4 & 86.6 \\
         COIN (ensemble)  &  {\bf86.2} & {\bf 87.0} \\
        \hline
    \end{tabular}
    \vspace{2mm}
    \caption{
    Experimental results on LCQMC.
    }
    \label{tab:lcqmc-result}
    \vspace{-4mm}
\end{table}

\begin{table}[t]
\fontsize{8}{10}\selectfont
 \setlength{\tabcolsep}{1.56mm}
  \centering
    \begin{tabular}{|l|c|r|}
        \hline
        {\bf Model} & {\bf Acc. (\%)} & {\bf Params}\\
        \hline
        BiMPM~\cite{wang2017bilateral} & 88.2 & 1.6M \\
        DIIN~\cite{gong2017natural} & 89.0 & 4.4M \\
        CAFE~\cite{tay-etal-2018-compare} & 88.7 & 4.7M \\
        OSOA-DFN~\cite{liu-etal-2019-original} & 89.0 & 10.0M \\
        RE2~\cite{yang-etal-2019-simple} & 89.2 & 2.8M \\
        ESAN~\cite{hu-etal-2020-enhanced} & 89.3 & 3.9M \\
        Enhanced-RCNN~\cite{peng2020enhanced} & 89.3 & 7.7M \\
        COIN (ours) &  {\bf89.4} & 6.5M   \\
        \hline
         BERT~\cite{devlin-etal-2019-bert} & 90.1 & 109.5M  \\
         SBERT~\cite{reimers-gurevych-2019-sentence} & 90.6 & 109.5M \\
         COIN (ensemble) &  {\bf90.7} & 32.5M \\
        \hline
    \end{tabular}
    \vspace{2mm}
    \caption{
    Experimental results on Quora dataset. 
    }
    \label{tab:quora-result}
    \vspace{-4mm}
\end{table}

\subsection{Experimental Results}\label{results}
We compare our model against recent prior work, including state-of-the-art neural models and BERT based methods. ESIM~\cite{chen-etal-2017-enhanced} and BiMPM~\cite{wang2017bilateral} are two strong sentence-interaction baselines. GMN~\cite{chen-etal-2020-neural-graph} is a neural graph matching network with multi-granular input information. DIIN~\cite{gong2017natural} extracts semantic features from the interaction space.
OSOA-DFN~\cite{liu-etal-2019-original} uses multiple original semantics-oriented attention, and
RE2~\cite{yang-etal-2019-simple} adopts richer features for alignment processes to improve the performance. ESAN~\cite{hu-etal-2020-enhanced} is a sentence-interaction model with gated feature augmentation. For pre-trained methods, we consider BERT~\cite{devlin-etal-2019-bert} and SBERT~\cite{reimers-gurevych-2019-sentence}. We also include ensemble results of our method where we consider the majority vote of the results given by 5 runs of the same model under different random parameter initialization.

Results on LCQMC are listed in Table~\ref{tab:lcqmc-result}. Our single model achieves better accuracy and F1-score than all non-pretrained baselines, and the results of COIN are fairly comparable to BERT despite \emph{not} being pretrained on any extrinsic data. In fact, our ensemble model (5 runs) even outperforms BERT.

The results on Quora are given in Table~\ref{tab:quora-result}. Our approach outperforms the non-pretrained baselines with 89.4\% test accuracy, and our ensemble model again achieves better results than BERT and SBERT with fewer parameters (32.5M vs.\ 109.5M). This confirms our model's ability to be applied in real-world scenarios that require less computational complexity and a smaller model footprint.

Overall, the above results on two question matching datasets reflect our model's effectiveness at capturing semantic interactions between the sentences and properly interring their relationship. In-depth analyses of the model's efficiency are given in the appendix.

\subsection{Model Analysis}\label{ablation}

\begin{table}[t]
\fontsize{9}{11}\selectfont
\vspace{-2mm}
 \setlength{\tabcolsep}{1.2mm}
  \centering
    \begin{tabular}{|l|cc|}
        \hline
        Model & Quora & LCQMC \\
         \hline
        original & 89.6 & 85.4   \\
        w/o context & 89.1 & 84.8  \\
        simple fusion & 88.8 & 85.2 \\
        w/o aggregat. & 89.2 & 84.9  \\
        simple pool & 89.4  & 85.2  \\
        \hline
    \end{tabular}
    \vspace{2mm}
    \caption{
    Ablation study on Quora and LCQMC dev set.
    } 
    \label{tab:ablation}
    \vspace{-2mm}
\end{table}

\smallskip
\noindent\textbf{Effect of Model Components.} 
In Table~\ref{tab:ablation}, we study the contribution of different model components. Without context in cross-attention, the accuracy decreases by 0.5 and 0.6 percentage points, respectively. This confirms that, by incorporating the context, our model can better capture sentence relationships in the alignments. We then replace the gate fusion with a simplified fusion layer, where we feed the concatenation of the two representations to a feed-forward network, observing a performance drop on both datasets. This shows the effectiveness of our context-aware interaction blocks.
We then remove the aggregation layer, finding that the accuracy decreases to $89.2\%$ and $84.9\%$. This confirms that the aggregation layer is useful to produce high-level representations for the final prediction. In the last ablation, we replace the multi-head pooling by max-pooling to produce the sentence vector, and the results decrease on both datasets.

\begin{figure}[t]
    \centering
    \vspace{-2mm}
    \includegraphics[scale=0.53]{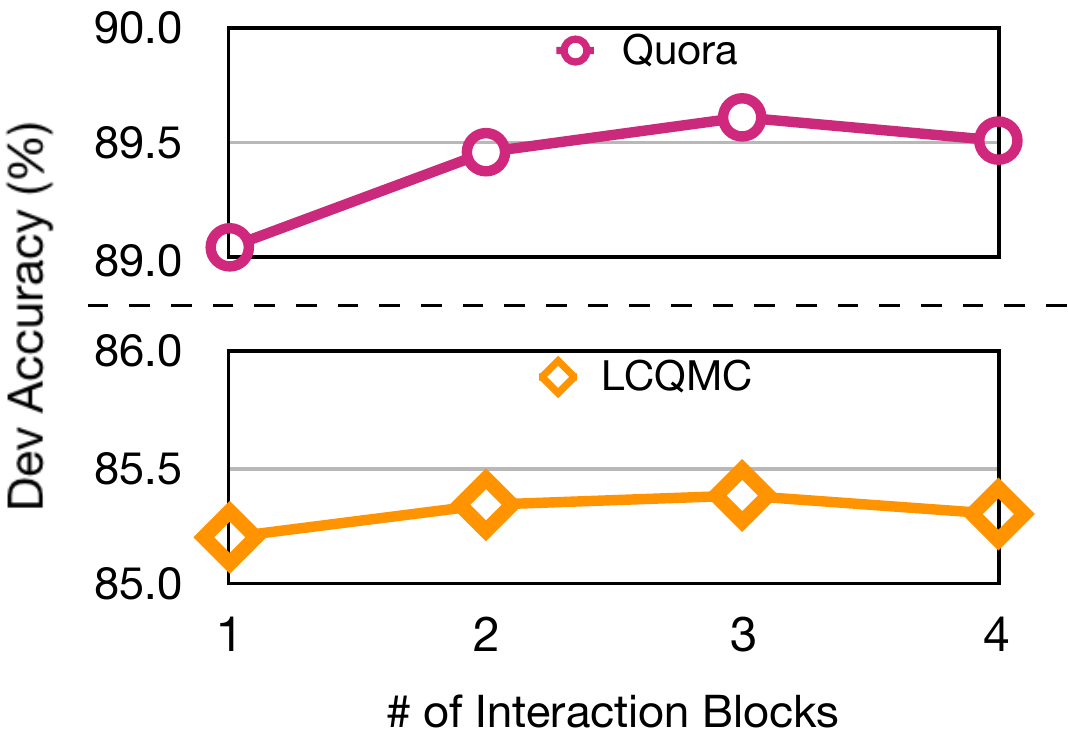}
    \vspace{-3mm}
    \captionof{figure}{ 
     Ablation study [Left] and effect of number of interaction blocks [Right] on Quora and LCQMC dev sets.
    }
    \label{fig:interaction_num_ablation}
    \vspace{-2mm}
\end{figure} 

\smallskip
\noindent\textbf{Effect of Interaction Block Depth.}
Figure~\ref{fig:interaction_num_ablation} plots the accuracy with varying numbers of interaction blocks. Evidently, a smaller number of interaction blocks may not suffice to fully capture the sentence relationships, and adding further such blocks may improve the model's ability to reason across the sequences and boost the performance. However, increasing the depth of interactions more than necessary harms the performance. Additionally, there is a trade-off between performance and efficiency since adding more interaction blocks increases the number of  parameters. For computational cost reasons, we use at most three interactions blocks in our experiments.

\smallskip
\noindent\textbf{Case Study.}

\begin{figure}[t]
    \centering
    \includegraphics[scale=0.32]{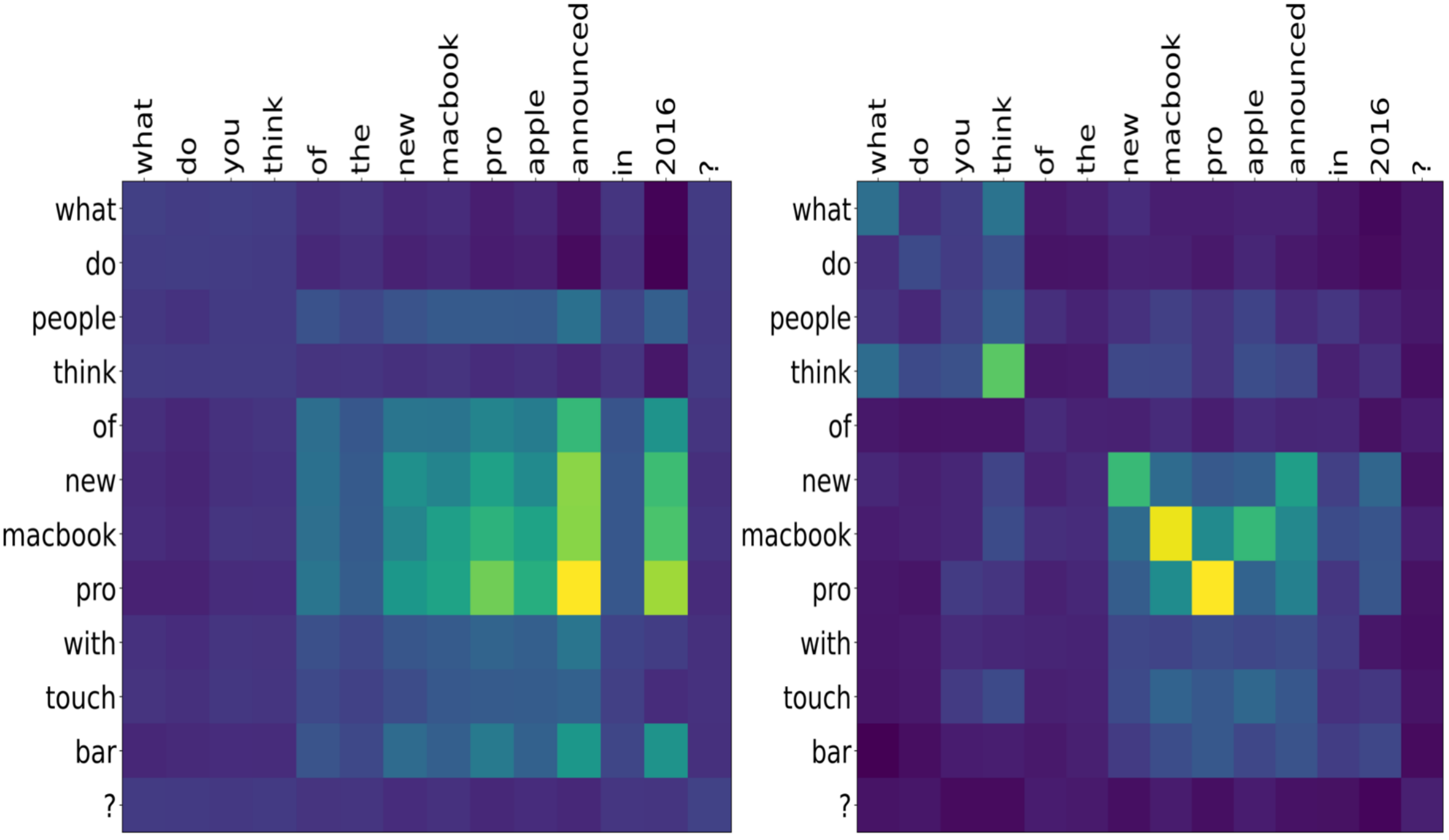}
    \vspace{-7mm}
    \captionof{figure}{ 
    Visualization of alignment in the first and third interactions. Lighter colors indicate higher values.
    }
    \label{fig:att-visual}
    \vspace{-4mm}
\end{figure}

We analyze the context-aware interaction results by visualizing the attention to show how the model learns aligned features at different levels of interaction in Figure~\ref{fig:att-visual}. We consider a sample from Quora with the target label \textsc{Duplicate}.

The left image shows the contextualized cross-attention in the first interaction block. Aided by the context, the model learns to correctly align the salient phrase ``\textit{new macbook pro}" across the inputs. 
The attention results in the third interaction block are visualized in the right image. As we can observe, the model refines the alignment results with a sharper distribution on the salient phrases than in the first interaction block, and the structured phrase ``\textit{what do you think of}" is also connected.  
The model thus predicts the relationship between the two sentences correctly. This corroborates our model's ability to gradually refine and adjust the attention scores in higher layers.

\section{Conclusion}
In this work, we propose a context-aware interaction network for question matching. We improve the cross-attention by incorporating contextual cues, and further leverage a gate fusion layer to flexibly integrate the aligned features. 
Experiments on two datasets validate the effectiveness of our architecture and show that accounting for the context enhances the original cross-attention.

\section*{Acknowledgements}
We thank the anonymous reviewers for their constructive suggestions.

\bibliography{emnlp2021}
\bibliographystyle{acl_natbib}

\appendix

\section{Experiment Details}
\noindent \textbf{Data Statistics.}
Statistics of the datasets are given in Table~\ref{tab:stats}. For LCQMC, we follow the same data split as in the original work~\cite{liu-etal-2018-lcqmc}, and for Quora we use the same split as \newcite{wang2017bilateral}.

\begin{table}[h]
\fontsize{8}{10}\selectfont
 \setlength{\tabcolsep}{1.1mm}
  \centering
    \begin{tabular}{|l|rrrc|}
        \hline
        {\bf Dataset} & {\bf Train} & {\bf Dev} & {\bf Test} & {\bf \# Classes} \\
        \hline
        \textsc{Quora} & 384K  & 10K & 10K & 2  \\
        \textsc{LCQMC} & 239K  & 9K & 13K & 2  \\
        \hline
    \end{tabular}
    \vspace{2mm}
    \caption{
    Statistics on the datasets for experiments.
    }
    \label{tab:stats}
    \vspace{-3mm}
\end{table}

\smallskip
\noindent \textbf{Preprocessing.}
We apply a hard cut-off of the sentence length on both datasets by cropping or padding.
Recent work has shown that character-based models typically outperform word-based models over Chinese NLP tasks~\cite{li-etal-2019-word-segmentation}, so we apply character-based modeling for LCQMC.
For Quora, we set the length as 32, and for LQCMC we set the length as 50. We mask the padding tokens during the experiments. 

\smallskip
\noindent \textbf{Embedding Details.}
For Quora, we use 300-dimensional GloVe CommonCrawl 840B word embeddings \cite{pennington-etal-2014-glove} and fix the weights during training. For LCQMC, following \newcite{li-etal-2019-word-segmentation}, we avoid word segmentation and instead use a randomly initialized character embedding matrix. We set the dimensionality of character embeddings to 200, and train the weights. For sentence preprocessing, we tokenize and lowercase all words. For efficiency and more generalizable results, we do not incorporate any additional lexical features in our experiments.

\smallskip
\noindent \textbf{Training Details.}
The kernel size is 3 for convolutional layers with padding. We apply 2 layers of convolutional encoder and 1 layer of convolutional aggregation in all experiments. We tune the dimensionality of the feed-forward layers from 150 to 300, and the number of interaction blocks from 2 to 4. The batch size is tuned from 32 to 128. Adam optimization is used with an initial learning rate of 0.001 and exponential decay. 
We apply ReLU~\cite{pmlr-v15-glorot11a} as the activation function in all feed-forward networks. To prevent over-fitting, dropout with a retention probability of 0.8 is applied.

Cross-entropy serves as the loss function during training. Adam optimization is used with an initial learning rate of 0.001, and $\beta_1$ is set as 0.9 and $\beta_2$ as 0.999 during training. Exponential decay is also applied. Moreover, we add L2 regularization, and set the threshold for gradient clipping as 5. We apply 3 context-aware interaction blocks for Quora, and 2 interaction blocks for LCQMC. We implement our model using TensorFlow~\cite{abadi2016tensorflow} and train the models on NVIDIA Tesla V100 GPUs and NVIDIA Tesla P4 GPUs. For BERT~\cite{devlin-etal-2019-bert}, we choose the BERT-base version (12 layers, 768 hidden dimensions and 12 attention heads), and fine-tune the model using the official implementation\footnote{https://github.com/google-research/bert}. The Chinese pre-trained BERT is adopted from https://huggingface.co/bert-base-chinese.
For SBERT~\cite{reimers-gurevych-2019-sentence}, we utilize the original implementation\footnote{https://github.com/UKPLab/sentence-transformers}, and add a softmax classifier on top of the output of the two Transformer networks as in the original paper.

\section{Model Efficiency}
\begin{table}[h]
\fontsize{8.5}{11}\selectfont
 \setlength{\tabcolsep}{1.3mm}
  \centering
    \begin{tabular}{lcc}
        \toprule
        \textbf{Models}  & {\bf parameter size} & {\bf time (s/batch)} \\
        \midrule
        COIN & 6.5M & 0.12 $\pm$ 0.03 \\
        BERT & 109.5M & 1.19 $\pm$ 0.06\\
        \bottomrule
    \end{tabular}
    \vspace{1mm}
    \caption{
    Parameter size and inference time for COIN and BERT on Quora dataset.
    }
    \label{tab:inference_time}
    \vspace{-5mm}
\end{table}

Pretrained language models such as BERT~\cite{devlin-etal-2019-bert} have drawn much attention for their substantial gains across a range of different natural language processing tasks. However, BERT is fairly demanding in terms of the computational requirements. For additional analysis, we compare our model efficiency with BERT-base on Quora. We set the sentence lengths as 32 (64 for BERT after concatenating the two sequences). Both models need to make predictions for a batch of 8 sentence pairs on a MacBook Pro with Intel Core i7 CPUs. For BERT, we add a linear layer on top of the \textsc{[CLS]} token for classification, as in the original paper. We report the average and the standard deviation of processing 1,000 batches.

As shown in Table~\ref{tab:inference_time}, COIN contains far fewer parameters than BERT and is much faster in terms of the CPU inference speed. Additionally, our single model produces comparable results to BERT on both Quora and LCQMC. This shows that our proposed method is effective at tackling text matching tasks with substantially fewer parameters and high computational efficiency.




\end{document}